\documentclass[letterpaper]{article} 
\usepackage{aaai20}  
\usepackage{times}  
\usepackage{helvet} 
\usepackage{courier}  
\usepackage[hyphens]{url}  
\usepackage{graphicx} 
\usepackage{graphicx}  
\usepackage{amsfonts}       
\usepackage{amsmath}
\usepackage{nicefrac}       
\usepackage{microtype}      

\title{Transfer Learning Robustness in Multi-Class Categorization by Fine-Tuning Pre-Trained Contextualized Language Models}
\author{Written by Xinyi Liu\textsuperscript{\rm 1*} and Artit Wangperawong\textsuperscript{\rm 2*}\\ 
\textsuperscript{\rm 1}University of Rochester, 
xinyi.liu1@ur.rochester.edu\\
\textsuperscript{\rm 2}U.S. Bank, 
artit.wangperawong@usbank.com 
}
 
\begin{document}

\maketitle

\begin{abstract}
This study compares the effectiveness and robustness of multi-class categorization of Amazon product data using transfer learning on pre-trained contextualized language models. Specifically, we fine-tuned BERT and XLNet, two bidirectional models that have achieved state-of-the-art performance on many natural language tasks and benchmarks, including text classification. While existing classification studies and benchmarks focus on binary targets, with the exception of ordinal ranking tasks, here we examine the robustness of such models as the number of classes grows from 1 to 20. Our experiments demonstrate an approximately linear decrease in performance metrics (i.e., precision, recall, $F_1$ score, and accuracy) with the number of class labels. BERT consistently outperforms XLNet using identical hyperparameters on the entire range of class label quantities for categorizing products based on their textual descriptions. BERT is also more affordable than XLNet in terms of the computational cost (i.e., time and memory) required for training. In all cases studied, the performance degradation rates were estimated to be 1\% per additional class label.
\end{abstract}

\section{Introduction}
\noindent Classification involves the prediction of one outcome from two or more possible discrete outcomes. A classification model predicts the probabilities of different possible outcomes of a categorically distributed dependent variable for a set of independent variables as input. Binary classification involves two possible outcomes, whereas multinomial (i.e., multi-class) classification involves three or more possible outcomes. Such models are based on the assumption of independence of irrelevant alternatives \cite{iia}, implying that the probability of preferring one class over another does not depend on the presence or absence of other irrelevant alternatives. For example, adding the class label \textit{Beauty} would not affect the model's relative output probabilities for classifying the example to be labeled under other categories. In practice, this assumption can be violated when a newly introduced class label correlates with any of the existing labels, e.g. \textit{Beauty} vs. \textit{Health and Personal Care}, as a given product may reasonably be multi-labeled. Since many interesting classification problems can involve a plethora of class labels for categorization, it is important to understand how the number of classes affects model performance.

Intuitively, one would expect a classification model's performance to decrease as the number of classes increases. After all, it would be more challenging even for humans to pick the right choice among more available options to categorize an item. Despite what we would imagine is a commonly occurring scenario, there is a dearth of studies or literature that closely examines the robustness of a certain model as the number of classes increases. 

\section{Background and Related Work}
Existing studies that compare the performance of different models in  multi-class categorization problems usually only examine a chosen couple high number of classes \cite{uglov,Chamasemani,Hasan}. With such few data points, we would not be confident in extrapolating a trend or judging the performance between different models and parameters. As we show in this work, \textbf{Model A} might outperform \textbf{Model B}  at a certain number of class labels but might underperform by adding or removing just one or two class labels (see Figs.~\ref{fig:1}~and~\ref{fig:2}). To conduct a comprehensive study, one would need machine learning models that can independently achieve high performance on each of the individual class labels. Applying transfer learning to state-of-the-art pre-trained models can be a feasible approach. Here we thus report on our experiments involving multi-class categorization by fine-tuning pre-trained contextualized language models to classify e-commerce products based on their textual descriptions on Amazon.

BERT and XLNet are two models that have recently achieved high performance on many natural language benchmarks, including text classification for positive and negative sentiments and ratings \cite{BERT,XLNet}. BERT-based models have lead the score boards for many NLP benchmark tasks since October 2018. More recently in June 2019, XLNet was shown to outperform BERT-related models on SQuAD, GLUE, RACE and 17 datasets, achieving record results on a total of 18 tasks. Only one month later in July 2019, an improved model called RoBERTa was demonstrated based on BERT with more training and removing an unnecessary training task (i.e., next-sentence prediction) \cite{RoBERTa}. With RoBERTa, the effect of data supply and hyperparameters were compared between XLNet and BERT, raising the question of whether one model is better than the other. In this study, we evaluated both models trained with fixed data supply and hyperparameters (e.g. learning rate, batch size, training steps etc.). The robustness of BERT and XLNet on multinomial classification was evaluated by categorizing textual product descriptions into product categories using Amazon product data \cite{Amazon}.

BERT stands for Bidirectional Encoder Representations from Transformers. This model learns deep bidirectional representations from unlabeled text by jointly conditioning on both left and right contexts in all layers. BERT's pre-training includes the Masked Language Model (MLM) where some of the tokens in the input are randomly masked for the model to predict the vocabulary id of the masked tokens given both left and right contexts. BERT also utilizes a next-sentence prediction task to train text-pair representations, which was demonstrated by RoBERTa to be unnecessary and disadvantageous for certain tasks. In this paper, we adopted the pre-trained BERT base model that includes 12 layers of a bidirectional transformer encoder \cite{Transformer} each with a hidden size of 768 units and 12 self-attention heads to comprise a total of 110 million parameters. The first token of each sequence inputted into the model is always a special token: [CLS]. The final hidden state of this token is treated as the final aggregate sequence representation of the sequence for the final classification. 

XLNet is a generalized autoregressive pre-training model. Instead of relying on masking the input to train the fused representation for the left and right contexts as in BERT, XLNet learns the bidirectional context by maximizing the expected log likelihood over all permutations of the factorization order. As the permutations can make the tokens from both the left and right contexts available at one side of the positions, XLNet overcomes the limitations of traditional autoregressive models that can be only trained unidirectionally to represent context and then concatenate them. It also avoids potential pretrain-finetune discrepancy caused by data corruption from masking tokens in BERT. Though permutated, XLNet encodes the positions to keep the positional information. XLNet adopts the recurrence mechanism and relative positional encoding scheme of Transformer-XL \cite{XL} to learn dependency beyond a fixed length and reparameterizes it to remove the ambiguity of the factorization order.

Given a set of target tokens $T$ and a set of non-target tokens $N=\textbf{x}\backslash T$, BERT and XLNet both maximize $\log p(T \mid N)$ but with different formulations:

\begin{equation}
    J_{\text{BERT}} = \sum_{x \in T} \log p(x \mid N),
\end{equation}
\begin{equation}
    J_{\text{XLNet}} = \sum_{x \in T} \log p(x \mid N \cup T_{<x}),
\end{equation}

\noindent where $T_{< x}$ denote tokens in $T$ that have a factorization order prior to $x$. We extended both pre-trained models by adding a softmax layer with the appropriate number of class outputs, $K$, to the final layer of the pre-trained models, resulting in output probabilities for the $j$\textsuperscript{th} class according to

\begin{equation}
    p(y=j \mid \textbf{q}) = \frac{e^{\textbf{q}^{\top} \textbf{w}_j }}{\sum_{k=1}^{K} e^{\textbf{q}^{\top} \textbf{w}_k }},
\end{equation}

\noindent where $K \in [1..20]$, \textbf{q} is the output from the final layer of the pre-trained model that is input into the softmax function, and \textbf{w} is the weight parameter to be trained. 

\begin{table}
    \centering
    \caption{Categories from Amazon products numbered in the order in which they are included in the models produced (Figs.~\ref{fig:2} and ~\ref{fig:3}).}
    \smallskip
     \begin{tabular}{c l}
    \hline
    \textbf{Order ($\boldsymbol{j}$)} & \textbf{Categories}\\ 
    \hline
1 & Musical instruments \\ \hline
2 & Baby \\ \hline
3 & Patio, Lawn and Garden \\ \hline
4 & Grocery and Gourmet Food \\ \hline
5 & Automotive \\ \hline
6 & Pet Supplies \\ \hline
7 & Office Products \\ \hline
8 & Beauty \\ \hline
9 & Tools and Home Improvement \\ \hline
10 & Toys and Games \\ \hline
11 & Health and Personal Care \\ \hline
12 & Cell Phones and Accessories \\ \hline
13 & Sports and Outdoors \\ \hline
14 & Kindle Store \\ \hline
15 & Home and Kitchen \\ \hline
16 & Clothing, Shoes, and Accessories \\ \hline
17 & CDs and Vinyl \\ \hline
18 & Movies and TV \\ \hline
19 & Electronics \\ \hline
20 & Books \\ \hline
    \end{tabular}
\label{table:1}
\end{table}

\section{Experimental Methods}
We fined-tuned the publicly available base models for BERT and XLNet (i.e., \textbf{BERT-Base} and \textbf{XLNet-Base}, respectively) with the above modifications for text classification on up to 20 different class labels. Both models have 12 transformer layers each with a hidden size of 768 units and 12 self-attention heads. 

For each additional class label, we added 5000 randomly sampled textual descriptions from the $j$\textsuperscript{th} category in the order shown in Table~\ref{table:1}. A random 10\% of all resulting samples were further held out for testing purposes and excluded from training. Although a product might reasonably fall under multiple categories and sub-categories, only the highest level label is used as the target. We only included descriptions with more than five characters to ensure sufficient textual information for the model inputs. 

For tokenization, we used BERT's pre-trained WordPiece tokenizer \cite{wordpiece} and XLNet's pre-trained SentencePiece tokenizer \cite{sentencepiece}. We used the uncased model for BERT and the cased model for XLNet according to what their respective authors reported to perform better. We developed the fine-tuned models using TensorFlow \cite{tensorflow}, and then trained it on an Nvidia Tesla T4 GPU for $n$ training steps, which is given by

\begin{equation}
\label{eq:training_steps}
    n = \frac{f \cdot K \cdot N_{samples}}{B} \cdot N_{epochs} ,
\end{equation}

\noindent where the hyperparameters $f$, $N_{samples}$, $N_{epochs}$, and $B$ are the train-test split ratio, number of samples per class, number of epochs, and mini-batch size, respectively, defined in Table~\ref{table:2}.

\begin{table}
    \centering
    \caption{Hyperparameter settings used to fine-tune BERT and XLNet pre-trained models.}
    \smallskip
     \begin{tabular}{l c}
    \hline
    \textbf{Hyperparameter} & \textbf{Value}\\ 
    \hline
Transformer layers ($L$) & 12 \\ \hline
Self-attention heads per layer ($A$) & 12 \\ \hline
Hidden size per layer ($H$) & 768 \\ \hline
Max sequence length ($L_{seq}$) & 128 \\ \hline
Train-test split ratio ($f$) & 0.9 \\ \hline
Samples per class ($N_{samples}$) & 5000 \\ \hline
Number of epochs ($N_{epochs}$) & 3 \\ \hline
Mini-batch size ($B$) & 32 \\ \hline
Learning rate ($\alpha$) & 2e-5 \\ \hline
Dropout rate ($p_{dropout}$) & 0.1 \\ \hline
    \end{tabular}
\label{table:2}
\end{table}

\begin{figure}[!hbt]
    \centering
    \includegraphics[width=0.95\linewidth]{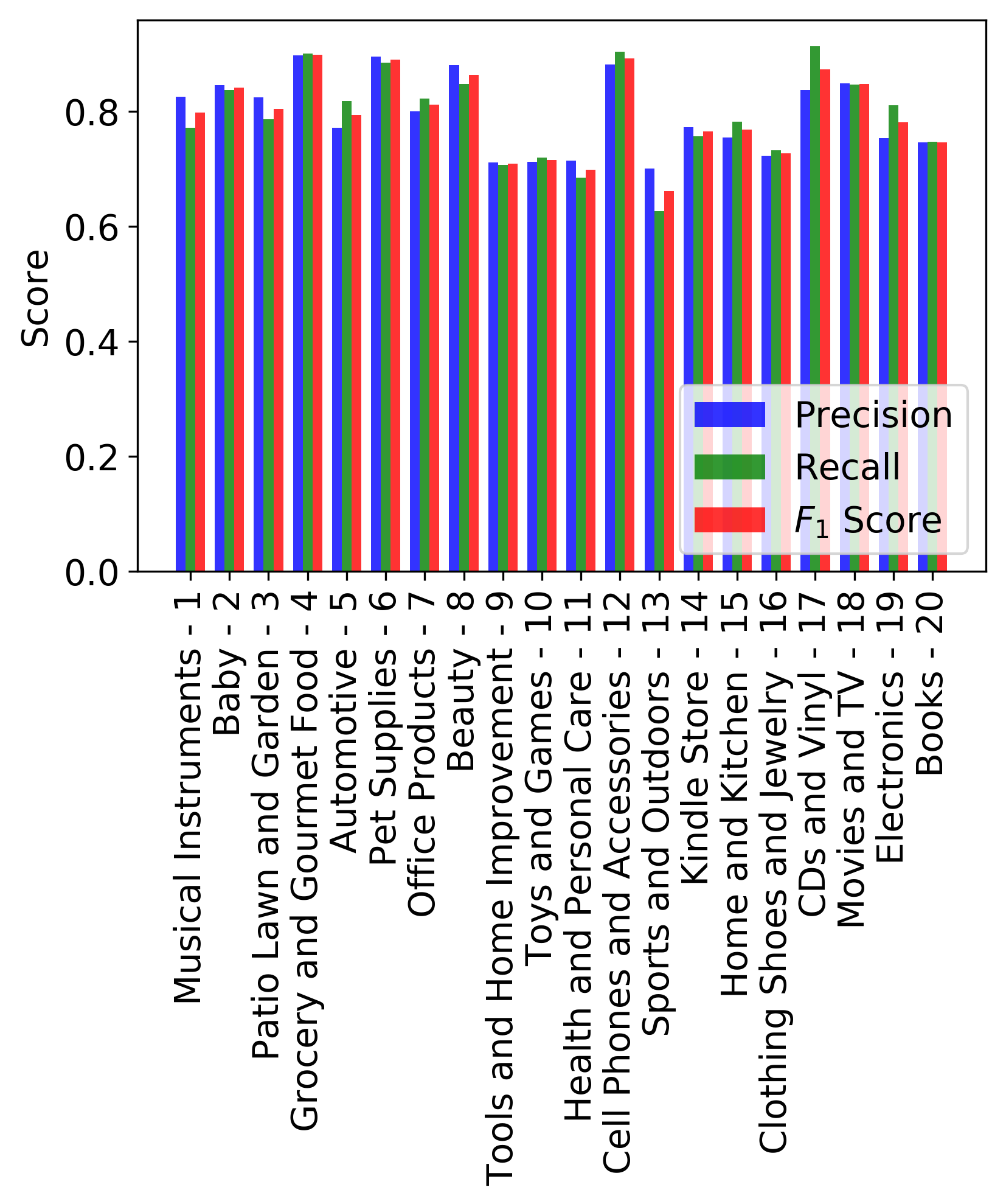}
    \caption{XLNet precisions, recalls, and $F_1$ scores of each category before averaging for $K=20$.}
    \label{fig:1}
\end{figure}

Having sampled balanced data from each class $j$ to avoid bias, we evaluated both models in terms of the macro-averaged precision, recall, $F_1$ score, and accuracy \cite{metrics} of the test set. To compute the macro-average, we first computed the metrics for each class. An example for $K=20$ is shown in Fig.~\ref{fig:1}. For class $j$, the precision, recall and $F_1$ score are calculated respectively as
\begin{equation}
    precision_{j} = \frac{TP_{j} }{TP_{j}+FP_{j}} ,
\end{equation}
\begin{equation}
    recall_{j} = \frac{TP_{j} }{TP_{j}+FN_{j}} ,
\end{equation}
\begin{equation}
    {F_1}_{j} = \frac{2\cdot precision_{j}\cdot recall_{j}}{precision_{j}+recall_{j}} ,
\end{equation}
where $TP_{j}$ denotes the number of true positive predictions, $FP_{j}$ denotes the number of false positive predictions and $FN_{j}$ denotes the number of false negative predictions for class $j$. We then calculated the average of the precisions, recalls and $F_1$ scores over all the classes as the macro-averaged precision, recall and $F_1$ score. The formulas with K total classes are respectively
\begin{equation}
    precision = \frac{\sum_{j=1}^{K} precision_j}{K} ,
\end{equation}
\begin{equation}
    recall = \frac{\sum_{j=1}^{K} recall_j}{K} ,
\end{equation}
\begin{equation}
    F_1 = \frac{\sum_{j=1}^{K} {F_1}_j}{K} ,
\end{equation}
\noindent and the accuracy of the test set is calculated as
\begin{equation}
    accuracy = \frac{\sum_{j=1}^{K} TP_j}{(1-f)\cdot K \cdot N_{samples} } .
\end{equation}

\section{Results and Discussion}

\begin{figure*}[!hbt]
    \centering
    \includegraphics[width=0.8\textwidth]{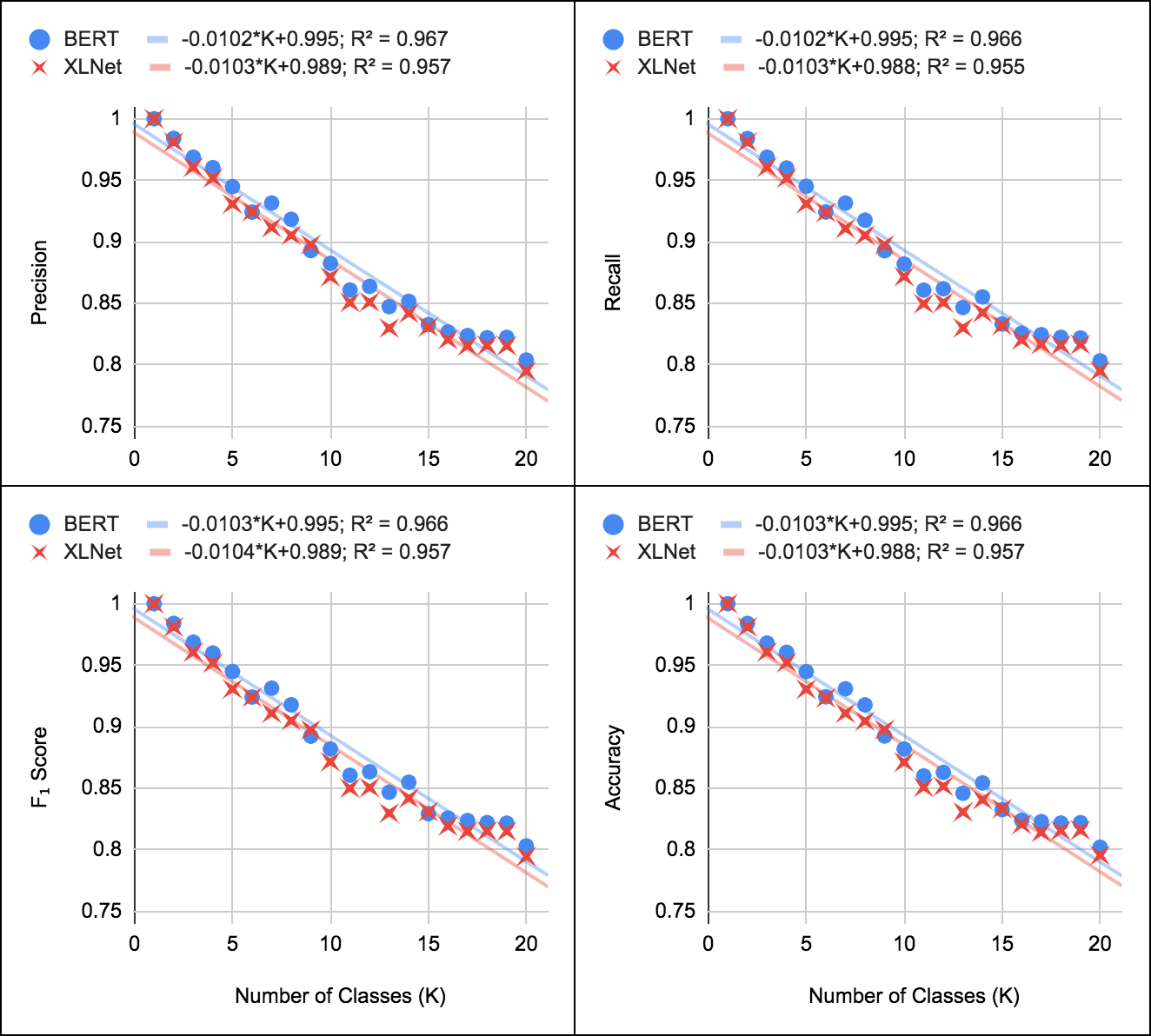}
    \caption{Macro-averaged precision, recall, $F_1$ score, and accuracy of the test set versus the number of classes $K$ for pre-trained \textbf{BERT-Base} and \textbf{XLNet-Base} language models fine-tuned for multi-class categorization. For completeness, the trivial case of a single class is included.}
    \label{fig:2}
\end{figure*}

All performance metrics on the test set decreases approximately linearly as the number of classes increases (Fig.~\ref{fig:2}). For completeness, the trivial case of a single class is included. The linearly decreasing trend suggests that the probability of incorrectly categorizing an item is proportional to the number of classes, i.e.
\begin{equation}
    p( y \neq j | x_j ) \propto K .
\end{equation}

\noindent By applying fitted lines using this relationship, we can therefore define a performance degradation rate of approximately 1\% per additional class for all the models studied here.

Since we used balanced data for each class, the macro-averaged recall should equal accuracy. However, due to rounding and floating point precision, negligible differences exist. It is also not surprising to see that the precisions and $F_1$ scores are similar to the accuracies. The linearly decreasing trend is preferable over an otherwise faster decreasing function, such as an exponential decay. However, it is possible that the initial linear trend can become non-linear with more classes.

Fitted lines result in coefficients of determination ($R^2$) close to 1, which indicate good fits in the range of interest. However, the dispersion of the data points appears to increase with the number of classes. Extrapolating performance would be therefore less accurate with higher number of classes. Deviations from the fitted line could be due to a variety of factors, including underfitting, overfitting, violating the assumption of independence of irrelevant alternatives as described above, and multi-label tendencies of a particular class.

Multi-label tendencies refers to an item reasonably having more than one label. Since this study is strictly multi-class categorization, the models have to choose one label for each item even if there is definition overlap between two or more class labels. For example, there could be items appropriately labeled as either \textit{Health and Personal Care} or \textit{Beauty}. We defer the topic of multi-label classification to future studies and will not discuss the issue further here.

\begin{figure*}[!hbt]
    \centering
    \includegraphics[width=0.8\textwidth]{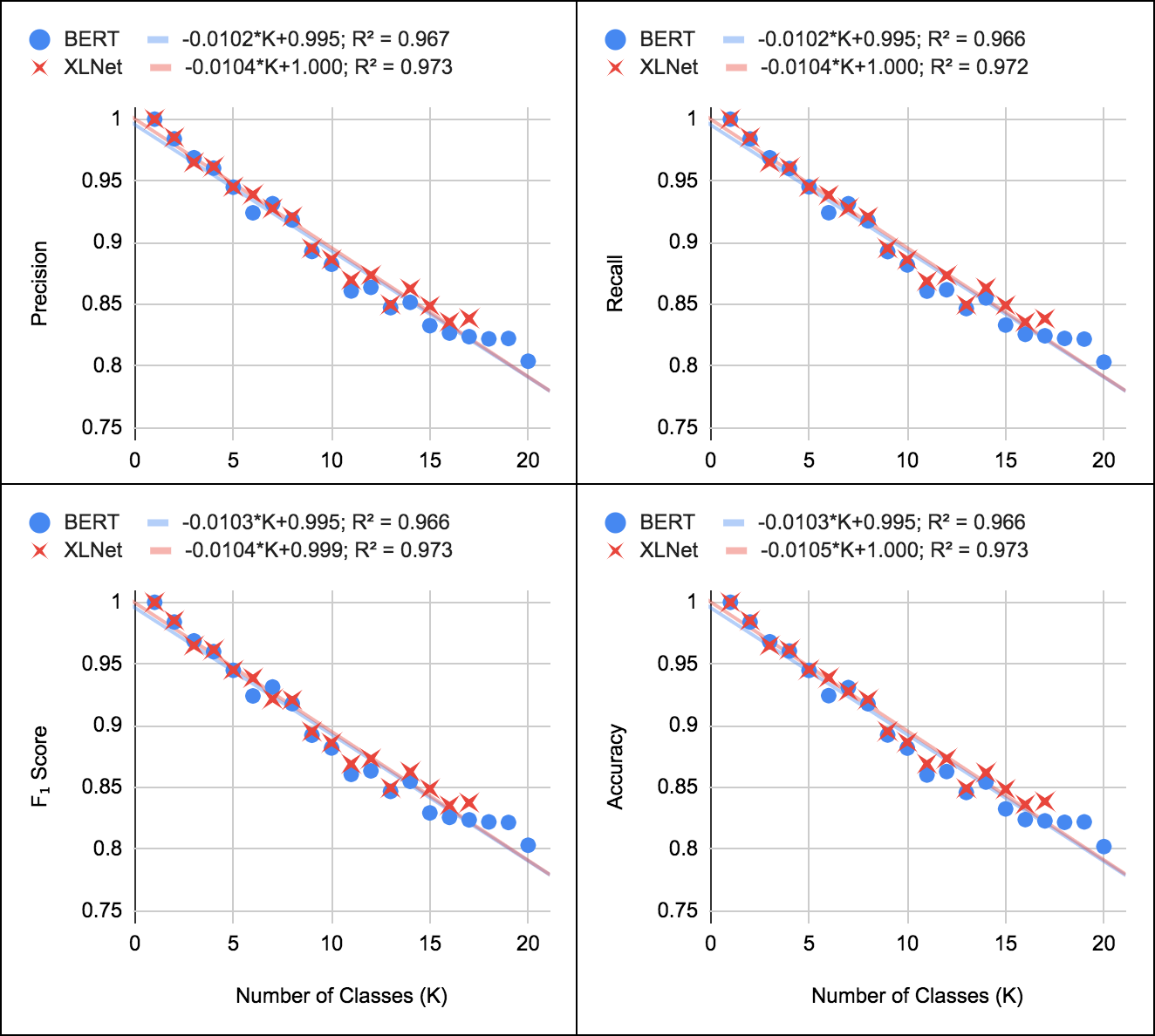}
    \caption{Macro-averaged precision, recall, $F_1$ score, and accuracy of the test set versus the number of classes $K$ for pre-trained \textbf{BERT-Base} and \textbf{XLNet-Large} language models fine-tuned for multi-class categorization. For completeness, the trivial case of a single class is included. Note that XLNet-Large data points were not available for all the number of classes due to computational resource limitations.}
    \label{fig:3}
\end{figure*}

Some classes perform more poorly than others, dragging down the macro-averaged metrics below the fitted line. As shown in Figs.~\ref{fig:2}~and~\ref{fig:3}, this is most evident when $K=11$ and $K=13$. A simple reason is that the respective latest included categories are more difficult to predict, i.e. \textit{Health and Personal Care} and \textit{Sports and Outdoors} (Fig.~\ref{fig:1}). Both BERT and XLNet exhibit the same trends, though BERT outperforms XLNet on all performance metrics over the entire range of class label quantities (Fig.~\ref{fig:2}). We show that a more performant model can reduce this discrepancy by reproducing the experiments described above on \textbf{XLNet-Large} (see Fig.~\ref{fig:3}). 

XLNet-Large is a much larger model than either XLNet-Base or BERT-Base, containing 24 transformer layers each with a hidden size of 1024 units and 16 self-attention heads. It also took approximately quadruple the time and triple the memory required for training XLNet-Base under the same experimental conditions described above. An exception was that we had to reduce the batch size to 8 to avoid memory limits. The number of training steps was maintained according to the formula described in Eqn.~\ref{eq:training_steps}. Despite our efforts, we were unable to obtain the results for all the number of classes due to computational resource limitations. Nevertheless, XLNet-Large is more robust in terms of performance with higher number of class labels. XLNet-Large is also better than BERT-Base in maintaining the linear trend; for example, see $K=6$ in Fig.~\ref{fig:3}. Even with the performance boost and discrepancy reduction offered by using this larger model, however, BERT-Base remains competitive over many values of $K$ examined.

\section{Conclusion and Future Work}

Experiments were conducted to understand the effect of including additional classes on multi-class categorization over a considerable range of class label quantities. We used identical hyperparameters to fairly compare BERT and XLNet models fine-tuned on Amazon product description data. BERT-Base outperformed XLNet-Base over the entire range studied with approximately 60\% less training time and 85\% less training memory. In order for XLNet to be competitive with BERT, we increased the number of parameters for the model by implementing XLNet-Large, quadrupling time and tripling memory requirements. However, BERT still remained competitive. In all cases, the models exhibited approximately linearly decreasing trends in performance metrics with the number of class labels. From the data collected in all the experiments, we estimate performance degradation rates of approximately 1\% per additional class.

To improve performance further, future work can explore the hyperparameter space. The number of unique training samples per class can be increased to the tens of thousands using the same Amazon product dataset. More computational resources can be applied. For instance, \textbf{BERT-Large} with whole-word masking, more training steps, and larger batch sizes can be used. Multi-modal fusion models can be developed by combining our fine-tuned models with image classifiers, which can also be based on transfer learning of state-of-the-art models \cite{multimodal}. Character-level models based on transformers can also be considered \cite{artitw}. In all cases, it would also be useful to study more than 20 classes to confirm if the linear trend persists. We have published code based on this study as open-source software called \textbf{Text2Class}, which is available on GitHub \cite{text2class_github} and the Python Package Index (PyPI) \cite{text2class_pypi}. 

\bibliographystyle{aaai}
\bibliography{citation.bib}

\end{document}